%% file: 0_main.tex
\documentclass[letterpaper, 10 pt, conference]{ieeeconf}  

\IEEEoverridecommandlockouts                              
\usepackage[utf8]{inputenc}
\usepackage[T1]{fontenc}

\usepackage{preamble}

\title{\LARGE \bf
A Topological Representation with Object–Path Graphs for Open-Vocabulary Instance Navigation}

\author{%
    Linwei Zheng$^{1}$,
    Daojie Peng$^{1}$,
    Bingtao Wang$^{1}$,
    Haoang Li$^{1}$,
    and Jun Ma$^{1,2}$
    \thanks{$^{1}$The Hong Kong University of Science and Technology (Guangzhou)}
    \thanks{$^{2}$The Hong Kong University of Science and Technology}
}

\begin{document}

\maketitle
\thispagestyle{empty}
\pagestyle{empty}

\begin{abstract}
\input{1_abstract}
\end{abstract}
\input{2_body}

\bibliographystyle{ieeetr}
\bibliography{IEEEabrv, references}

\end{document}

%% file: 1_abstract.tex
Vision-language navigation requires embodied agents to navigate environments using natural language instructions and visual observations. Existing approaches typically decompose navigation into sequential language-guided decisions or rely on online exploration without prior environmental knowledge. Scene graph representations offer compact semantic memory but remain decoupled from downstream navigation, which still depends on dense metric maps. To close this gap, we propose an object--path graph that unifies open-vocabulary semantic reasoning with topological navigation. The proposed representation jointly supports semantic grounding, graph-based localization, and navigation within a single lightweight topological framework. Building on this graph, we introduce a navigation strategy that combines global path planning with local inter-node execution through lightweight node localization and semantic visual servoing, enabling navigation directly over the graph without dense metric reconstruction. Experiments on HM3D and Replica demonstrate competitive performance in open-vocabulary object grounding through the proposed hierarchical graph structure, while achieving effective navigation performance. Real-world robot experiments further validate the practicality of the proposed framework.

%% file: 2_body.tex
\section{INTRODUCTION}

Vision-language navigation (VLN) aims to enable embodied agents to navigate environments using natural language instructions together with visual observations. Unlike traditional robotic navigation systems that depend on manually engineered rules or dense metric maps, VLN requires agents to jointly perform multimodal perception, semantic reasoning, spatial understanding, and sequential decision-making ~\cite{zhan2024mc, wen2025tinyvla, peng2025lovon}. These capabilities are essential for broad applications in household robotics, autonomous systems, emergency response, and human-robot interaction.

Recent advances in large vision-language models (VLMs) and large language models (LLMs) have significantly accelerated progress in VLN research ~\cite{zhou2024navgpt}. By leveraging pretrained semantic knowledge and language reasoning, modern VLN systems can 
interpret complex instructions and perform navigation with improved generalization. Existing methods commonly transform visual observations into semantic representations and use language-guided reasoning to predict actions, subgoals, or exploration strategies. Open-vocabulary perception further enables agents to recognize unseen object categories and reason about semantic relationships within the environment.

\begin{figure}[t]
    \centering
    \includegraphics[width=0.48\textwidth]{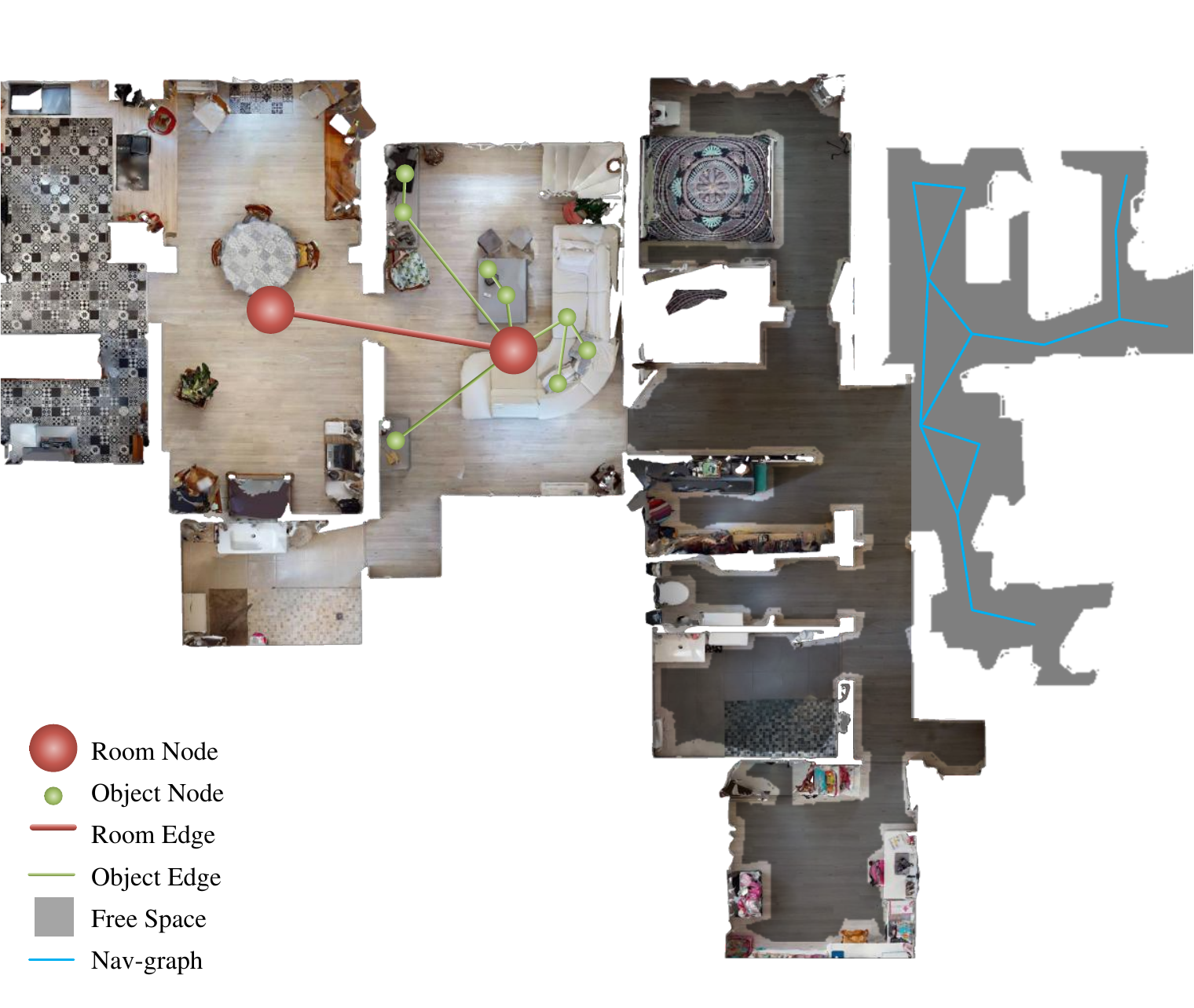}
    \caption{The object-path graph consists of two essential topological layers: an object-level graph encoding semantic entities and their relations, and a path-level graph capturing navigable connectivity between spatial nodes. }
    \label{fig:layer}
\vspace{-1em} \end{figure}

Despite recent progress, current VLN systems still face two fundamental limitations. Most approaches either rely on sequential language-guided decision making~\cite{visionandlanguage2018anderson, rajvanshi2024saynav}, where navigation is decomposed into step-by-step actions, or operate without prior environmental knowledge~\cite{sgnav2024yin, vlfm2024yokoyama}, requiring online exploration to locate the target. Consequently, navigation over long distances is prone to accumulated errors, redundant exploration, and inefficient trajectories.
Scene graph representations~\cite{conceptgraphs2024gu, werby2024hierarchical, visual2023huanga}   have recently emerged as a compact semantic memory for embodied agents, providing structured object relationships and supporting language reasoning. However, they are primarily used for high-level semantic understanding, while downstream navigation still depends on dense metric maps or continuous global localization.

To bridge this gap, we propose an object--path graph representation that unifies semantic understanding and topological navigation. As shown in Fig.~\ref{fig:layer}, the overall framework constructs an object graph for open-vocabulary semantic reasoning together with a path graph encoding navigable connectivity. Building upon this representation, we introduce a graph-based navigation strategy that combines global path planning with local inter-node execution, enabling topological navigation directly over the graph through lightweight node localization and semantic visual servoing.

The main contributions of this work are summarized as follows:

\begin{itemize}
\item We propose an object--path graph representation that unifies open-vocabulary semantic understanding and topological navigation within a single graph framework.

\item We introduce a graph-based navigation strategy that combines global planning with local inter-node execution, enabling navigation directly over the constructed graph without requiring dense metric maps.

\item We demonstrate the effectiveness of the proposed representation for multiple VLN tasks, including object grounding, node initialization, topological routing, and inter-node navigation, validating the feasibility of graph-centric navigation.
\end{itemize}

\section{Related Work}


Recent advances in VLMs have enabled zero-shot navigation through direct reasoning over visual observations. PIVOT~\cite{nasiriany2024pivot} formulates navigation as sequential visual reasoning, where VLMs select actions from observed frames. Other works introduce reusable spatial memories to improve long-horizon navigation. Value-map-based methods, such as VLFM~\cite{vlfm2024yokoyama} and WMNav~\cite{nie2025wmnav}, rank exploration regions according to task relevance, while semantic map-based approaches including ESC~\cite{zhou2023esc}, L3MVN~\cite{yu2023l3mvn}, and OpenFMNav~\cite{kuang2024openfmnav} integrate object semantics and spatial structures for viewpoint selection. Nevertheless, many of these representations are task-specific or require additional metric planning modules for navigation.
More recently, graph-based representations have been explored as unified memories for semantic reasoning and navigation. SG-NAV~\cite{sgnav2024yin} and SayNav~\cite{rajvanshi2024saynav} construct hierarchical scene graphs to support goal localization and reasoning, while MC-GPT~\cite{zhan2024mc} builds a topological memory over sampled viewpoints for long-horizon navigation. Related approaches such as ETPNav~\cite{etpnav2025an} and OVL-Map~\cite{ovlmap2025wen} further incorporate structured graph-like memories to support LLM-driven navigation policies and action selection.

These methods primarily treat graphs as intermediate memory structures for high-level reasoning, while low-level navigation relies on step-by-step instruction decisions, introducing additional overhead. In contrast, another line of work focuses on constructing explicit 3D scene representations for open-vocabulary understanding.
ConceptGraphs~\cite{conceptgraphs2024gu} builds open-vocabulary scene graphs from 3D observations, while VLMaps~\cite{visual2023huanga} constructs visual-language maps for language-guided navigation, which a follow-up work extends with acoustic information~\cite{multimodal2025huanga}. HOV-SG~\cite{werby2024hierarchical} further introduces hierarchical open-vocabulary scene graphs with room-level organization. Recent works such as OpenIN~\cite{openin2025tang} and DualMap~\cite{dualmap2025jiang} extend semantic mapping to dynamic settings. However, these approaches primarily focus on semantic representation and scene understanding, and do not directly formulate navigation over the graph structure itself. Our work bridges this gap by enabling global planning and topology-only navigation directly on the constructed graph.

\section{Methodology}
\begin{figure}[t]
	\centering

\includegraphics[width=0.48\textwidth]{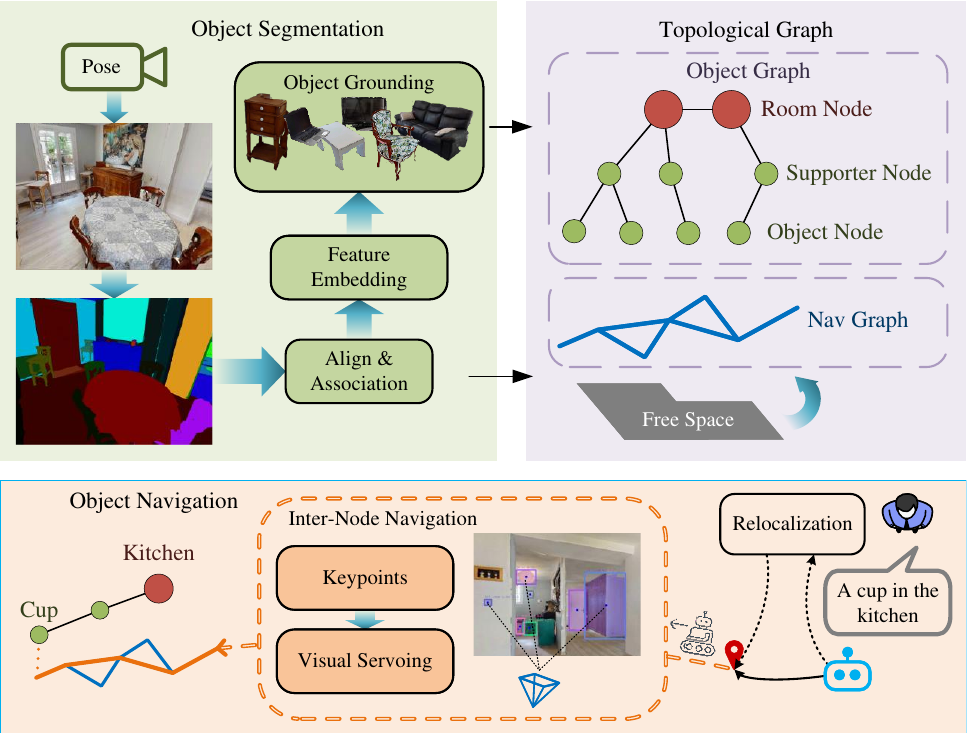}	

\caption{Overview of the proposed framework. The pipeline consists of four key components: Open-Vocabulary Semantic Segmentation for extracting object-level observations from RGB-D inputs, Object Scene Graph construction for encoding objects and their semantic relations, Traversability Navigation Graph for modeling navigable free-space connectivity, and Topological Navigation for performing goal-driven inter-node navigation using the constructed object–path graph representation.}
	\label{fig:overview} 
\vspace{-1em} \end{figure}

This section presents the proposed object--path graph framework for topological vision-language navigation. As shown in Fig.~\ref{fig:overview}, the framework first constructs an open-vocabulary object representation from RGB-D observations (Section~\ref{sec:object_mapping}), builds object and traversability graphs for semantic reasoning and spatial connectivity (Sections~\ref{sec:object_graph} and~\ref{sec:path_graph}), and finally performs graph-based navigation through node initialization and visual servoing (Section~\ref{sec:topological_navigation}).

\subsection{Open-Vocabulary 3D Object Mapping} \label{sec:object_mapping}
The objective of the open-vocabulary mapping module is to construct a compact semantic representation of the environment from RGB-D observations and pose measurements. Then we group geometrically and semantically consistent points into object-level segments. This significantly reduces representation redundancy while naturally producing object-centric entities that serve as the nodes of the proposed object graph. 

Given a sequence of RGB-D frames, we first apply the Segment Anything Model (SAM) to generate class-agnostic 2D segmentation masks. Using the corresponding depth maps, masked pixels are projected into 3D space to obtain local point-cloud segments. With odometry information, all segments are transformed into a shared global coordinate frame.
To maintain a consistent object representation, newly observed segments are associated with existing segments using geometric overlap and semantic similarity. The overlap between segments \(S_m\) and \(S_n\) is defined as \(O(m,n)=a(S_m,S_n)/\max(n(S_m),n(S_n))\), where \(a(S_m,S_n)\) denotes neighboring point correspondences within a distance threshold. Semantic similarity is measured by the cosine similarity of CLIP features, \(C(m,n)=f_m^\top f_n/(\|f_m\|\|f_n\|)\). Segments are merged only when both measures exceed predefined thresholds.

After association, the object feature is updated through weighted fusion:
\begin{equation}
    f=\sum_i w_i f_i,\qquad
    w_i=\frac{n(S_i)}{\sum_j n(S_j)},
\end{equation}
where \(n(S_i)\) denotes the number of retained points after downsampling. This weighting assigns greater importance to observations with stronger geometric evidence, producing a stable object-level semantic representation.

After processing the image sequence, VLMs generate object descriptions from informative multi-view image crops. The resulting captions are subsequently refined by an LLM into a single coherent object-level description for each object.

\subsection{Object Scene Graph}\label{sec:object_graph}
Recent approaches have explored room-level decomposition from structural observations~\cite{werby2024hierarchical}. Following this paradigm, we partition the environment into disjoint room regions using the projected structural layout and an Euclidean Distance Field (EDF). After segmentation, each room is represented as a room node. To assign semantic labels, we collect the detected objects contained within each room and prompt an LLM to generate a concise room description, resulting in an open-vocabulary semantic label (e.g., \emph{kitchen}, \emph{bedroom}, or \emph{office}).

Inspired by~\cite{openin2025tang}, we categorize detected objects within each room into \emph{supporter objects} and \emph{regular objects}.
Supporters correspond to large structural objects such as beds, desks, cabinets, and sofas that naturally organize the surrounding scene, while regular objects denote smaller movable items such as books, cups, and keyboards. Specifically, this categorization is determined based on the estimated volume of each object's 3D bounding box.  Each supporter is connected directly to its parent room node. Regular objects are associated with one or multiple nearby supporters according to their spatial proximity, allowing objects located on shared surfaces or between neighboring furniture to maintain multiple supporting connections. In addition, every object is assigned an LLM-generated caption describing its semantic attributes.

To capture semantic interactions beyond the hierarchical structure, we establish candidate edges between spatially adjacent objects using 3D proximity. For each candidate pair, object captions together with their relative spatial configurations are provided to an LLM, which predicts their semantic relationship (e.g., \emph{on top of}, \emph{inside}, or \emph{next to}). The inferred relation is stored as the edge attribute, replacing purely geometric adjacency with semantically meaningful connectivity.

Finally, adjacent room nodes are connected through traversable doorways to preserve inter-room topology. The resulting open-vocabulary scene graph \(G_O=(V_O,E_O)\) encodes room hierarchy, supporting structures, object-level semantic relations, and room connectivity. Within this graph, supporter and regular object nodes maintain object-level states for downstream grounding and navigation, while room nodes provide hierarchical semantic context.

\subsection{Traversability Navigation Graph}\label{sec:path_graph}

Navigation nodes are generated based on changes in surrounding views, where view similarity is measured by the ratio between the number of shared visible objects and the total number of observed objects from two viewpoints.

The initial set of navigation nodes is obtained from the recorded robot poses. A new node is added when the surrounding view differs significantly from the previous node, allowing the graph to preserve important transitions along the trajectory. Doorway regions between rooms are explicitly inserted as additional nodes to maintain inter-room connectivity. For free-space regions that are not sufficiently covered by the original trajectory, we randomly sample candidate positions and group them according to view similarity. The mean position of each group is then selected as an additional navigation node.

Edges are established between two nodes when the straight-line path connecting them is collision-free according to the reconstructed free-space representation. Since sparse node sampling may result in disconnected graph components, additional nodes are inserted between neighboring components to establish connectivity. The resulting traversability graph \(G_P(V_P,E_P)\) captures the topological connectivity of the environment and provides a compact representation for global planning and inter-node navigation.

\subsection{Object Navigation}\label{sec:topological_navigation}
\subsubsection{Map Representation}

Instead of relying on dense point cloud or occupancy grids, we represent the environment using a sparse object-grounded topological graph. The map is defined as
\begin{equation}
G_O = (V_O^{room} \cup V_O^{obj}, E_O),
\end{equation}
where \(V_O^{room}\) denotes room nodes, \(V_O^{obj}\) denotes object nodes including supporters and regular objects, and \(E_O\) represents semantic and hierarchical relationships among nodes.

Object nodes \(v_i \in V_O^{obj}\) store a structured state representation:
\begin{equation}
v_i = \{ \mathbf{p}_i, \mathbf{f}_i, c_i, d_i\},
\end{equation}
where \(\mathbf{p}_i = (x_i, y_i, z_i)\) is the 3D position of the object centroid, \(\mathbf{f}_i \in \mathbb{R}^d\) is the semantic feature embedding extracted from VLMs, and \(c_i\) is the object category or open-vocabulary caption. \(d_i\) is the volume of the object. This formulation jointly encodes geometric connectivity and semantic relationships in a unified graph structure, providing a compact representation for downstream planning and navigation.

\subsubsection{Node Initialization}

We formulate initial localization as a coarse node retrieval problem over the constructed object--path graph. The robot first performs a short-range exploration to acquire sufficient surrounding observations for constructing a panoramic semantic descriptor.

To enable efficient node retrieval, we employ a lightweight semantic retrieval framework\cite{place2025zheng}. Each navigation node is represented by a compact semantic panoramic descriptor. Instead of encoding geometric laser signatures, the descriptor represents the surrounding semantic observations of each node. Specifically, visible objects are projected into a discretized panoramic representation with \(N\) angular bins, where each bin aggregates the semantic features of the corresponding visible objects. The resulting panoramic descriptor \(\mathbf{s}\) is made rotation-invariant by considering circular shifts during descriptor matching, enabling robust retrieval under unknown heading alignment.

Given the query descriptor \(\mathbf{s}_{\mathrm{query}}\), the starting node is determined through comparing the cosine similarity:
\begin{equation}
    v^* = \arg\max_{v_i \in V_P}
    \mathrm{cos}(\mathbf{s}_{\mathrm{query}},\mathbf{s}_i),
\end{equation}
where \(\mathbf{s}_i\) denotes the semantic panoramic descriptor stored at navigation node \(v_i\).

\subsubsection{Topological Navigation}

Given the current navigation node \(v^*\) obtained from node initialization and the target node \(v_T\) identified through object grounding, a global path
\(
P=(v^*=v_0,\,v_1,\ldots,v_M=v_T)
\)
is computed using a shortest-path search (e.g., Dijkstra's algorithm) over the traversability graph \(G_P(V_P,E_P)\). The resulting path defines an ordered sequence of navigation nodes connecting the start and target locations. Inter-node navigation then executes this route as a visual servoing process, where each step transitions from the current node \(v_k\) to the next node \(v_{k+1}\). Rather than relying on explicit metric localization or dense trajectory optimization, the agent directly aligns its current observation with the panoramic snapshot stored at \(v_{k+1}\).

During map construction, a panoramic snapshot is created for every navigation node. Each panoramic image is processed using an open-vocabulary segmentation model to obtain object-level observations. For every segmented object, the image centroid is approximated by
\begin{equation}
    \mathbf{c}=\frac{1}{|S|}\sum_{(u,v)\in S}(u,v),
\end{equation}
where \(S\) denotes the object mask. Each keypoint is represented by its center, object label, and semantic feature embedding. To avoid ambiguous correspondences, only distinctive keypoints are retained. Specifically, a keypoint is preserved if its label is unique within the observation, or, for repeated labels, its feature embedding differs sufficiently from the other detections according to a cosine similarity threshold.

During navigation, the current observation is processed in the same manner, and correspondences with the target-node snapshot are established by matching object labels followed by cosine similarity verification between their semantic feature embeddings.

Given the current observation and target snapshot, navigation is formulated as a visual servoing process that aligns their object arrangements. Let \(\mathbf{k}_i^c\) and \(\mathbf{k}_i^g\) denote matched keypoints in the current and goal observations, respectively. Instead of minimizing pixel displacement, we measure the relative angular discrepancy between consecutive keypoints, which is more robust to camera translation and viewpoint variation. Since the robot operates in planar motion, keypoints are projected onto the XY-plane, and the geometric residual is defined as:

\begin{equation}
\mathbf{e}_i =
(\theta_{i+1}^{c} - \theta_i^{c}) -
(\theta_{i+1}^{g} - \theta_i^{g}),
\end{equation}

where \(\theta_i^c\) and \(\theta_i^g\) represent the horizontal angular positions of the \(i\)-th matched keypoint in the current and goal observations, respectively. The residual measures the difference in relative object arrangement between the two views and provides the basis for generating the navigation control command.
The overall alignment objective is defined as:
\begin{equation}
\mathcal{E} = \sum_i w_i \|\mathbf{e}_i\|,
\end{equation}
where \(w_i\) denotes the confidence weight derived from semantic similarity. The aggregated residual is then mapped to robot motion commands through the visual servoing controller:
\begin{equation}
\mathbf{v}^t = f(\{\mathbf{e}_i^t\}) \rightarrow (v_x^t,v_y^t).
\end{equation}

\begin{figure}[t]
    \centering
    \includegraphics[width=0.3\textwidth]{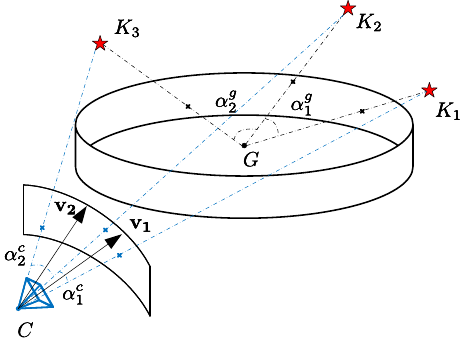}
    \caption{The robot at current position \(C\) observes a panoramic representation at the goal point \(G\), from which keypoints \(K_i\) are detected. The control objective is to compute a motion command that aligns the robot toward the goal configuration by minimizing the angular and geometric discrepancy between the current observation and the goal-view keypoint structure.}
    \label{fig:3k}
\vspace{-1em} \end{figure}

To illustrate the geometric intuition behind the proposed controller, we first consider a simplified case with three matched keypoints, as shown in Fig.~\ref{fig:3k}. The robot motion is determined by comparing the relative angular differences between the current and target observations.

If we define the angular alignment error as \(e_i=\alpha_i^{c}-\alpha_i^{g}\), where \(\alpha_i=\theta_{i+1}-\theta_i\) represents the relative angle between consecutive keypoints, the control objective is to reduce the temporal evolution of this error:
\begin{equation}
\dot{\mathbf{e}}_i=\dot{\alpha}_i^{c}.
\end{equation}
The relationship between the angular error dynamics and robot motion can be derived geometrically. Since \(\alpha_1\) and \(\alpha_2\) are independent, it suffices to analyze a single angular sector. We consider motion along an arbitrary direction \(\mathbf{v}_i\) within the sector formed by \(\overrightarrow{CK}_i\) and \(\overrightarrow{CK}_{i+1}\). The motion direction partitions the angular sector \(\alpha_i\) into two sub-angles, denoted by \(\alpha_{ii}^{c}\) and \(\alpha_{ii+1}^{c}\). Using trigonometric geometry together with the small-angle approximation, the error dynamics are obtained as
\begin{equation}
\dot{\mathbf{e}}_i
=
\left(
\frac{\sin(\alpha_{ii}^{c})}{|CK_i|}
+
\frac{\sin(\alpha_{ii+1}^{c})}{|CK_{i+1}|}
\right)
\operatorname{sgn}\!\left(\alpha_i^{c}-\alpha_i^{g}\right)
\mathbf{v}_i.
\end{equation}
Treating \(\mathbf{v}_1\) and \(\mathbf{v}_2\) as basis directions, we can enforce a negative error dynamics \(\dot{\mathbf{e}} < 0\), which drives the error vector \([\mathbf{e}_1, \mathbf{e}_2]^T \rightarrow \mathbf{0}\). This ensures that the robot is guided to the goal node.

Within the visual servoing framework, the control input is defined as a proportional mapping from the alignment error to motion commands. The motion along each bisector direction is decomposed onto adjacent edge vectors, 
leading to the following local velocity formulation:
\begin{equation}
\mathbf{v}_i =
-\left(\overrightarrow{CK}_i + \overrightarrow{CK}_{i+1}\right)
\left(\alpha_i^{c}-\alpha_i^{g}\right)
\cos\left(\frac{\alpha_i^{c}}{2}\right),
\end{equation}
where \(\overrightarrow{CK}_i\) here denotes the unit vector pointing toward the \(i\)-th keypoint. The overall control input is obtained by summing the contributions from all angular sectors,
\[
\mathbf{u}=\sum_i \mathbf{v}_i.
\]
The modulation term \(\cos(\alpha_i^{c}/2)\) serves two purposes. First, its sign changes when \(\alpha_i^{c}>\pi\), automatically reversing the control direction to maintain convergence. Second, its magnitude increases as \(\alpha_i^{c}\) decreases, providing a larger control gain when the observed angular configuration deviates significantly from the target, thereby providing enough excitation during the early stage of navigation while naturally reducing the control effort near the goal.

By stacking all edge contributions, the global motion can be expressed in matrix form as
\begin{equation}
\begin{bmatrix}
v_x \\
v_y
\end{bmatrix}
=
-\mathbf{A}\mathbf{B}\boldsymbol {\alpha},
\end{equation}
where \(\mathbf{A} = [\mathbf{I}, \mathbf{I}, \ldots, \mathbf{I}]\), \(\mathbf{B}\) is a block-diagonal matrix constructed from edge vectors \(\overrightarrow{CK}_i\) and \(\boldsymbol{\alpha}\) collects the angular alignment errors with cosine modulation.
This formulation compactly couples panoramic angular errors with the underlying spatial configuration of the navigation graph. Therefore, the overall controller can be expressed in the following weighted linear form:
\begin{equation}
\mathbf{u} = K \, \mathbf{w} \mathbf{e},
\end{equation}
where \(K\) is the induced gain matrix determined by the geometric decomposition, \(\mathbf{e}\) denotes the panoramic angular alignment error, and \(\mathbf{w}\) is a diagonal weight matrix derived from semantic similarity.

The inter-node navigation process is terminated when the alignment error satisfies \(
\|\mathbf{e}\| < \epsilon,
\)
where \(\epsilon\) is a predefined threshold.
Although the system only provides linear velocity commands \([v_x, v_y]\), the control vector can be further mapped to unicycle control inputs \([v, \omega]\) where a gain \(k\) adjusts the turning aggressiveness to produce smoother robot motion:
\begin{equation}
v = \sqrt{v_x^2 + v_y^2}, \quad
\omega = k \cdot \mathrm{atan2}(v_y, v_x).
\end{equation}
By transforming node-to-node navigation into a semantic visual alignment problem, the proposed controller enables the robot to follow topological connections through perceptual feedback. The resulting framework provides an efficient alternative to metric navigation, requiring only compact node representations for graph traversal.

\section{Experiment}

We evaluate the proposed topological mapping framework on goal-driven vision-language navigation tasks. The experiments are designed to analyze three aspects: (i) object query grounding capability using semantic and relational cues, (ii) starting node initialization through graph-based relocalization using the constructed object-level topological graph, and (iii) navigation performance over the inferred graph structure when the target object is correctly identified as a goal node.

\subsection{Object Query on the Graph Map}

We evaluate the proposed object--path graph on open-vocabulary object queries involving semantic, relational, and instance-level descriptions. Given a natural language query, an LLM extracts the target object together with optional supporter and room information. Retrieval is then performed hierarchically from room to supporter to target object, falling back to object-level search when higher-level constraints are unavailable.

We compare against representative open-vocabulary scene representations, including ConceptGraphs~\cite{conceptgraphs2024gu}, VLMaps~\cite{visual2023huanga}, and HOV-SG~\cite{werby2024hierarchical}, on the Habitat-Matterport 3D (HM3D)~\cite{habitatmatterport2021ramakrishnan} and Replica~\cite{replica19arxiv} datasets. All methods are evaluated under identical object query settings, where the agent retrieves the target object instance from a natural language query.

\begin{table}
\centering
\caption{Structural representation comparison across different methods.}
\label{tab:structure_support}
\begin{tabular}{lcc}
\hline
Method & Room Structure & Built-in Nav-graph\\
\hline
\textbf{ConceptGraphs~\cite{conceptgraphs2024gu}}& $\times$ & $-$\\
\textbf{VLMaps}~\cite{visual2023huanga}& $\times$ & $-$\\
\textbf{HOV-SG}~\cite{werby2024hierarchical}& $\checkmark$ & $\checkmark$ \\
\textbf{Ours} & $\checkmark$ & $\checkmark$\\
\hline
\end{tabular}
\vspace{-1em}  \end{table}

\begin{figure}
    \centering
    \includegraphics[width=0.5\textwidth]{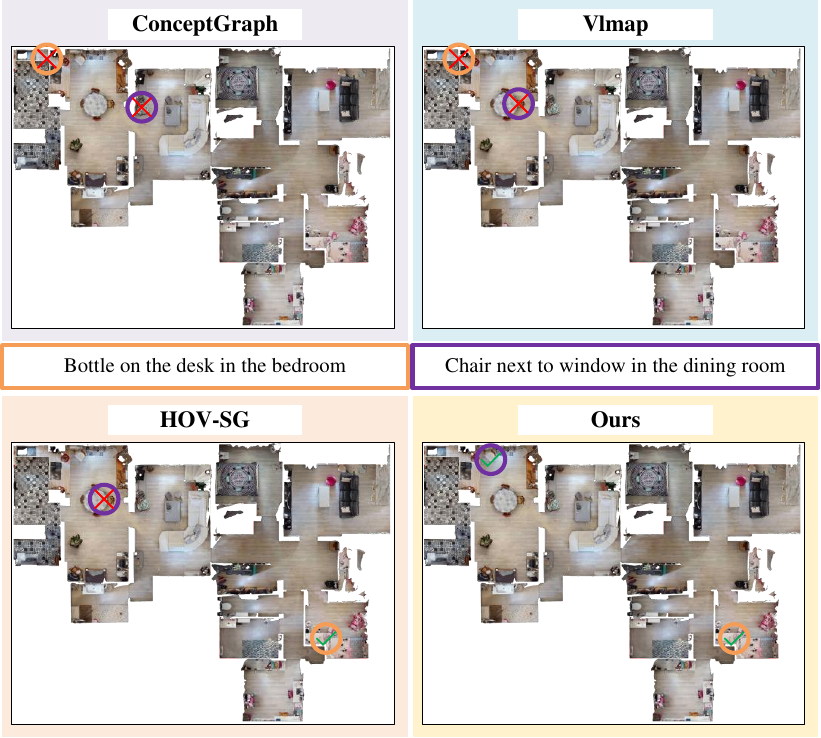}
    \caption{By jointly leveraging object–relation structure and room identification, our method achieves more accurate instance-level retrieval compared to competing approaches.}
    \label{fig:showcase}

\vspace{-1em} \end{figure}

As shown in Table~\ref{tab:structure_support}, ConceptGraphs and VLMaps do not maintain an explicit navigation graph for route planning. Instead, they rely on dense spatial representations, such as costmaps or occupancy grids, to perform geometric planning. HOV-SG adopts a Voronoi graph, also requiring global metric localization.
Object query performance is evaluated using Success Rate (SR). We evaluate object grounding with 120 queries across four scenes, covering object-only, relation-based, and room-aware descriptions. Table~\ref{tab:obj_query_sr} shows that our method consistently improves retrieval, especially for relational queries, highlighting the benefit of hierarchical semantic modeling. As illustrated in Fig.~\ref{fig:showcase}, this effectively resolves relational queries such as ``a book on top of a desk in the living room.''
We further evaluate object navigation using the retrieved graph node as the navigation goal. As shown in Table~\ref{tab:spl_table}, the proposed object--path graph achieves the highest average SPL, demonstrating that accurate object grounding naturally translates into efficient navigation.

\begin{table}[t]
\centering
\caption{Object query SR performance.}
\label{tab:obj_query_sr}
\begin{tabular}{lcccc}
\hline
Method & Object & Relation & Room+Relation & Avg. \\
\hline
ConceptGraphs~\cite{conceptgraphs2024gu} & 0.58 & 0.63 & 0.60 & 0.60 \\
VLMaps~\cite{visual2023huanga} & 0.60 & 0.70 & 0.75 & 0.68 \\
HOV-SG~\cite{werby2024hierarchical} & 0.63 & 0.75 & 0.85& 0.74\\
\textbf{Ours} & 0.68 & 0.88 & 0.93 & 0.83 \\
\hline
\end{tabular}
\vspace{-1em}
\end{table}

\begin{table}[t]
\centering
\caption{Object navigation performance in terms of SPL from 10 different start points.}
\label{tab:spl_table}
\begin{tabular}{lcccccc}
\hline
Method & Obj-1 & Obj-2 & Obj-3 & Obj-4 & Obj-5 & Avg. SPL \\
\hline
VLMaps~\cite{visual2023huanga} & 0.21 & 0.26 & 0.18 & 0.29 & 0.23 & 0.234\\
HOV-SG~\cite{werby2024hierarchical} & 0.34 & 0.38 & 0.31 & \textbf{0.46}& 0.36 & 0.37\\
\textbf{Ours} & \textbf{0.40} & \textbf{0.43} & \textbf{0.38} & 0.41& \textbf{0.42} & \textbf{0.408}\\
\hline
\end{tabular}
\vspace{-1em}  \end{table}

\subsection{Scene Graph Node Retrieval}

We evaluate the proposed node initialization strategy by retrieving the current navigation node from a short observation sequence. SceneGraphLoc~\cite{miao2024scenegraphloc} is one of the few methods addressing localization on scene graph representations by learning a cross-modal embedding between scene graphs and visual observations.

Since the original SceneGraphLoc constructs scene-level subgraphs that are too coarse for navigation, we instead generate subgraphs centered at navigation nodes while preserving its graph matching pipeline, making the comparison consistent with topological navigation.

As shown in Table~\ref{tab:retrieval_scenegraphloc}, the proposed method achieves comparable localization performance to SceneGraphLoc while using a simpler retrieval framework. These results demonstrate that navigation nodes equipped with compact semantic descriptors provide sufficient discriminative information for reliable topological node initialization.

\begin{table}[t]
\centering
\caption{Retrieval performance (Recall@K) comparison across scenes.}
\label{tab:retrieval_scenegraphloc}
\begin{tabular}{llcc}
\hline
Scene & Metric & SceneGraphLoc & Ours \\
\hline
Scene-1 & R@1 & \textbf{0.86}& 0.78 
\\
        & R@3 & 0.9
& \textbf{0.91}\\
        & R@5 & \textbf{0.97}& 0.94 \\
\hline
Scene-2 & R@1 & \textbf{0.80}& 0.72 
\\
        & R@3 & \textbf{0.93}& 0.87 
\\
        & R@5 & \textbf{0.96}& 0.92 \\
\hline
Scene-3 & R@1 & 0.81& \textbf{0.88}\\
        & R@3 & 0.92 & \textbf{0.96}\\
        & R@5 & \textbf{0.98}& 0.95\\
\hline
Scene-4 & R@1 & \textbf{0.75}& 0.74\\
        & R@3 & 0.89 & \textbf{0.90}\\
        & R@5 & 0.93 & \textbf{0.97}\\
\hline
\end{tabular}
\vspace{-1em}  \end{table}
\subsection{ Edge-Navigation Performance Evaluation}

We evaluate the proposed inter-node navigation strategy by measuring the accuracy of reaching target navigation nodes. Since the controller aligns the current observation with the semantic snapshot stored at the target node, its performance depends on both the available visual observations and the semantic information retained in the snapshot.

For each transition, the agent moves from the current node toward the target node using the proposed visual servoing strategy. The translation error is defined as the Euclidean translation distance:
\begin{equation}
    e_{pos}=||\mathbf{p}_r-\mathbf{p}_n||_2,
\end{equation}
where $\mathbf{p}_r$ and $\mathbf{p}_n$ denote the robot position and target node position.

\begin{figure}[t]
    \centering
    \includegraphics[width=0.5\textwidth]{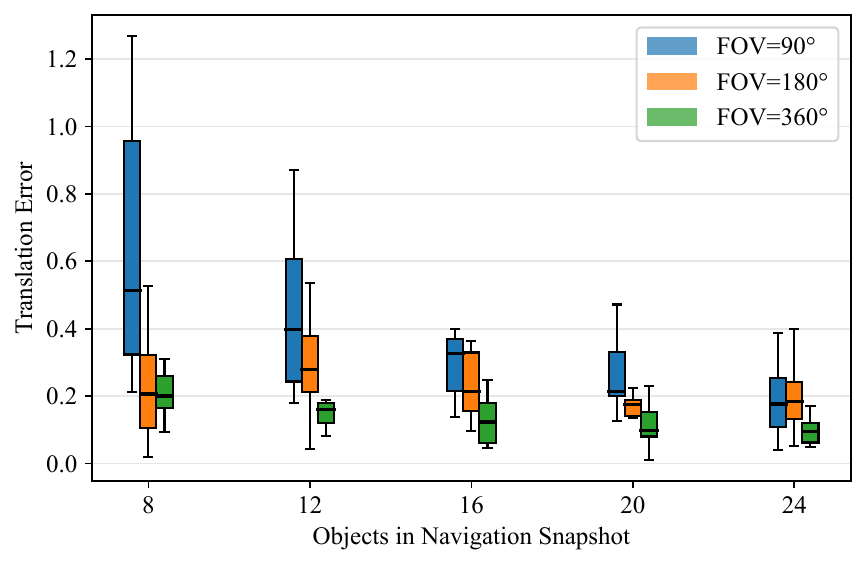}
    \caption{Translation error under different visual servoing configurations. The number of stored objects controls the semantic richness of each navigation-node snapshot, while different camera fields of view determine the available observations during inter-node navigation.}
    \label{fig:navigation_error}
\vspace{-1em} \end{figure}

We further analyze the influence of the camera field of view during navigation and the number of objects retained in each navigation-node snapshot. Fig.~\ref{fig:navigation_error} summarizes the resulting translation errors under different setups. Increasing the number of stored objects consistently improves convergence accuracy by providing richer semantic correspondences for visual alignment. When insufficient objects are retained (e.g., fewer than eight keypoints), narrow fields of view (90°) may cause insufficient correspondence and lead to off-route navigation. Panoramic observations further reduce the variance of the translation error, particularly when only a limited number of objects are available. Unless otherwise specified, all subsequent simulation experiments adopt panoramic observations during inter-node navigation.

We also report two metrics for evaluation of navigation. Navigation Success Rate (NSR) measures the percentage of node transitions successfully completed within the convergence threshold. The Path Length Ratio (PLR) is defined as the ratio between the executed trajectory length and the shortest path length on the navigation graph, evaluating the efficiency of the generated trajectory.

Table~\ref{tab:navigation_performance} summarizes the navigation performance across different HM3D scenes. The results demonstrate that the proposed topological navigation framework can reliably complete node-level navigation while maintaining efficient traversal over the constructed path graph.

\begin{table}[t]
\centering
\caption{Navigation performance.}
\label{tab:navigation_performance}
\begin{tabular}{lcc}
\hline
Scene & NSR $\uparrow$ & PLR $\downarrow$ \\
\hline
Scene-1 & 0.92 
& 1.18 
\\
Scene-2 & 0.89 
& 1.25 
\\
Scene-3 & 0.94 
& 1.15 
\\
Scene-4 & 0.87 
& 1.32 
\\
\hline
Average & 0.91 & 1.23 \\
\hline
\end{tabular}
\vspace{-1em}  \end{table}

The proposed framework achieves reliable node-level navigation across different scenes, indicating that the visual servoing strategy can effectively guide the agent through the constructed path graph. The high NSR demonstrates that the object-guided topological representation provides sufficient spatial constraints for accurate node transitions without requiring dense geometric localization.

The Path Length Ratio remains higher than the theoretical shortest path due to the nature of the current node-based navigation strategy. Since each transition is independently performed through visual servoing toward the next relay node, the trajectory does not explicitly optimize global path smoothness or continuous motion between consecutive nodes. Although this introduces additional intermediate movements and occasional stopping behavior at relay nodes, it provides a simple and robust mechanism for graph-based navigation. Future work will investigate continuous trajectory optimization over the topological graph to further improve navigation efficiency.

\begin{table}[t]
\centering
\caption{Ablation study on graph design and feature weighting strategy}
\label{tab:ablation_all}
\begin{tabular}{llcc}
\hline
Component & Setting & NSR $\uparrow$ & SPL $\uparrow$ \\
\hline
Graph & w/o room & 0.83 & 0.38 \\
& w/o supporter & 0.77& 0.32\\
& w/o semantic relation & 0.81 & 0.36 \\
\hline
Weighting& No weighting & 0.82 & 0.36 \\
\hline
Full model & with weighting& \textbf{0.88} & \textbf{0.42} \\
\hline
\end{tabular}
\vspace{-1em}  \end{table}

\subsection{Ablation Study}

We conduct ablation experiments to evaluate the contribution of different components in the proposed object--path graph representation and visual servoing strategy. Results are summarized in Table~\ref{tab:ablation_all}.

\subsubsection{Effect of Graph Structure}

We analyze the contribution of different structural components in the object graph, including room-level organization, supporter hierarchy, semantic relations, and spatial relations. Removing room-level information (\textit{w/o room}) degrades performance due to the loss of contextual constraints. Removing the supporter hierarchy (\textit{w/o supporter}) weakens object organization by eliminating the structural relationship between large supporting objects and smaller movable objects. Removing semantic relations (\textit{w/o semantic relation}) further reduces object discrimination, especially among visually similar objects.

These results demonstrate that hierarchical organization and relational reasoning are important for constructing an effective semantic representation for navigation.

\subsubsection{Effect of Feature Weighting Strategy}

We further evaluate the influence of feature weighting during visual servoing. Without weighting, all matched keypoints contribute equally to the control command, while similarity-based weighting emphasizes more reliable correspondences.

The results show that feature weighting improves both NSR and SPL by reducing the impact of ambiguous matches and producing more stable navigation commands.

\subsection{Real-World Navigation Validation}
To validate the transferability of the proposed topological edge navigation strategy, we conduct real-world experiments using a wheeled mobile robot equipped with a Jetson onboard computer and an ORBBEC Gemini 336L RGB-D camera. The experiment focuses on evaluating the execution of inter-node transitions, where the robot aligns its current observation with the semantic snapshot associated with the target navigation node.

\begin{figure}[t]
    \centering

    \begin{subfigure}{1\linewidth}
        \centering
        \includegraphics[width=\linewidth]{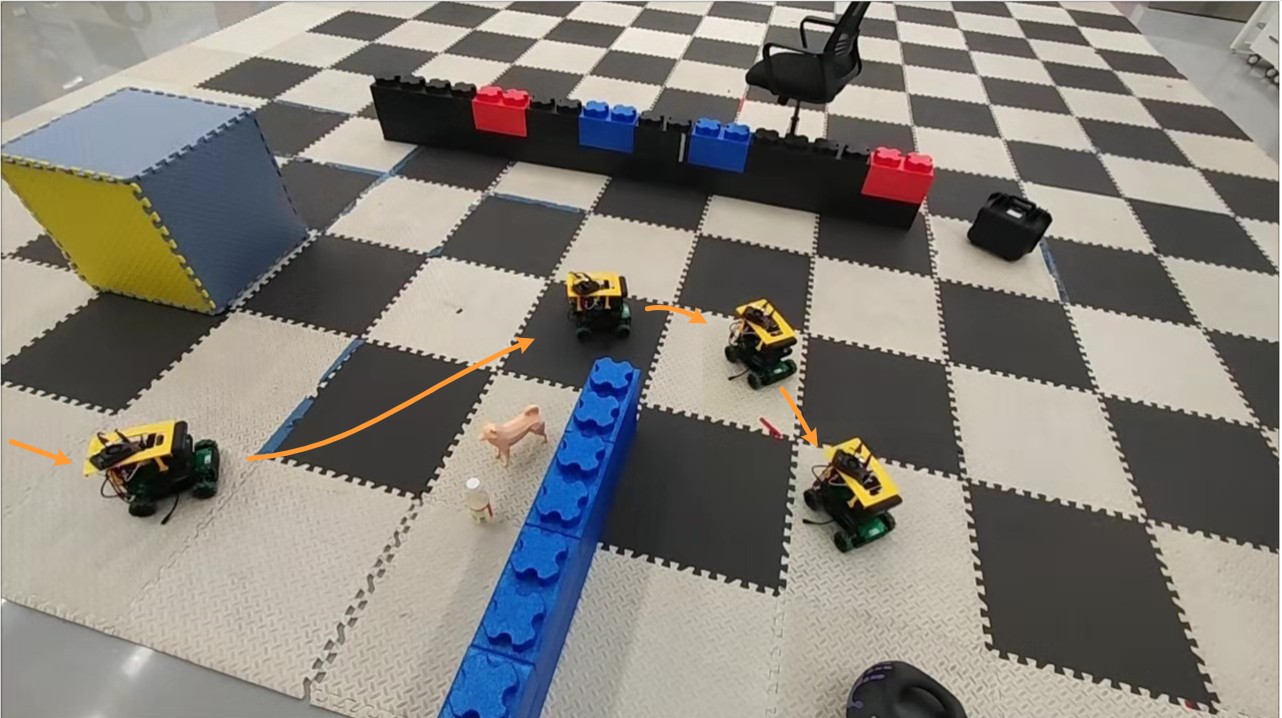}
       \caption{}
    \end{subfigure}

    \vspace{0.5em}

    \begin{subfigure}{0.49\linewidth}
        \centering
        \includegraphics[width=\linewidth]{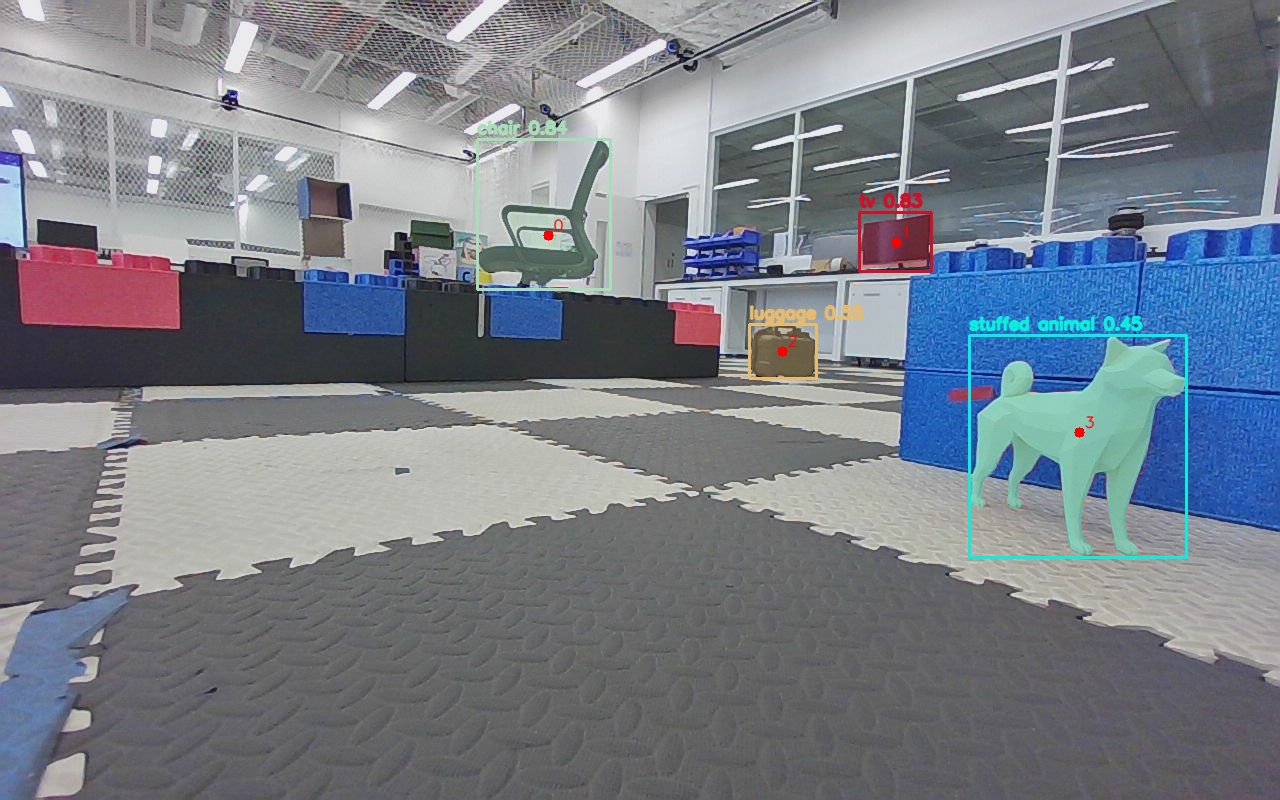}
      \caption{}
    \end{subfigure}
    \hfill
    \begin{subfigure}{0.49\linewidth}
        \centering
        \includegraphics[width=\linewidth]{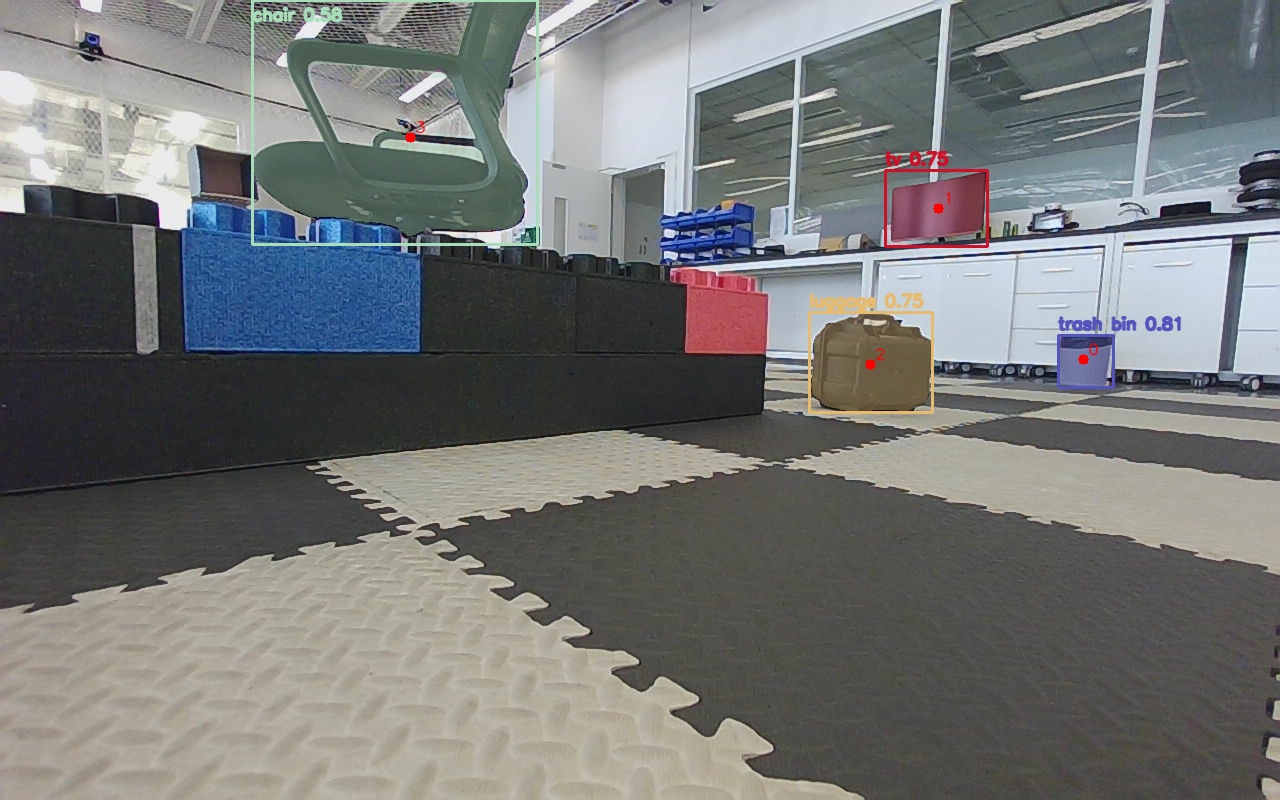}
          \caption{}
    \end{subfigure}

    \caption{Real-world validation of topological edge navigation. After establishing the topological path to the target object, the robot traverses graph edges by aligning current observations with the stored semantic representation of the next navigation node. 
(a) Navigation path between consecutive nodes.  (b) Current observation. (c) Observation at reaching point, where the observation aligns with the stored node snapshot}
    \label{fig:realworld}
\vspace{-1em} \end{figure}

The robot is instructed to traverse a sequence of navigation nodes toward a target location. During each transition, object observations are extracted from the RGB-D input and matched against the stored semantic snapshot of the next navigation node to estimate the control command. The experiment demonstrates that the proposed navigation strategy can reliably execute graph edges directly from sparse semantic node representations without requiring dense metric trajectory optimization. Fig.~\ref{fig:realworld} shows an example of a successful transition toward a target node. The robot progressively reduces the visual alignment error and converges to the desired location

\section{Conclusion}

The proposed framework constructs an object graph for semantic reasoning and a path graph for spatial traversal, providing a unified representation for VLN tasks including object grounding and navigation. This work demonstrates the feasibility of performing navigation directly on a topological map without requiring dense metric reconstruction or learned navigation policies. However, the current servo-based inter-node navigation does not explicitly optimize global path efficiency. Extending the framework to dynamic environments remains an important direction toward real-world deployment.